\documentclass{article}

     \usepackage[final]{neurips_2019}

\usepackage[utf8]{inputenc} % allow utf-8 input
\usepackage[T1]{fontenc}    % use 8-bit T1 fonts
\usepackage{hyperref}       % hyperlinks
\usepackage{url}            % simple URL typesetting
\usepackage{booktabs}       % professional-quality tables
\usepackage{amsfonts}       % blackboard math symbols
\usepackage{nicefrac}       % compact symbols for 1/2, etc.
\usepackage{microtype}      % microtypography
\usepackage{textcomp}
\usepackage{subfigure}
\usepackage{graphicx}
\usepackage{times}
\usepackage{helvet}
\usepackage{courier}
\usepackage{amsmath}
\usepackage{nccmath}
\graphicspath{ {./images/} }
\def\Nat{{\rm I\kern\pIR N}}

\def\argmin{\mathop{\rm arg\,min}}

\def\vec0{{\boldsymbol{0}}}
\def\vecb{{\boldsymbol{b}}}

\def\vecv{{\boldsymbol{v}}}
\def\vecw{{\boldsymbol{w}}}

\newcommand{\beq}{\begin{equation}}
\newcommand{\eeq}{\end{equation}}
\newcommand{\beqa}{\begin{eqnarray}}
\newcommand{\eeqa}{\end{eqnarray}}
\newcommand{\beqan}{\begin{eqnarray*}}
\newcommand{\eeqan}{\end{eqnarray*}}
\newcommand{\ben}{\begin{eqnarray*}}
\newcommand{\een}{\end{eqnarray*}}

\def\vecw{{\boldsymbol{\bf w}}}

\usepackage{color}      % microtypography
\usepackage{comment}

\title{Overcoming Catastrophic Interference in \\Online Reinforcement Learning with \\Dynamic Self-Organizing Maps}

\author{%
  Yat Long Lo\\
  Department of Computer Science\\
  University of Hong Kong\\
  \texttt{loyat@ualberta.ca} \\
  \And 
   Sina Ghiassian\\
  Department of Computer Science\\
  University of Alberta\\
  \texttt{ghiassia@ualberta.ca} \\
  }

\begin{document}

\maketitle
\begin{abstract}
Using neural networks in the reinforcement learning (RL) framework has achieved notable successes. Yet, neural networks tend to forget what they learned in the past, especially when they learn online and fully incrementally, a setting in which the weights are updated after each sample is received and the sample is then discarded. Under this setting, an update can lead to overly global generalization by changing too many weights. The global generalization interferes with what was previously learned and deteriorates performance, a phenomenon known as catastrophic interference. Many previous works use mechanisms such as experience replay (ER) buffers to mitigate interference by performing minibatch updates, ensuring the data distribution is approximately independent-and-identically-distributed (i.i.d.). But using ER would become infeasible in terms of memory as problem complexity increases. Thus, it is crucial to look for more memory-efficient alternatives. Interference can be averted if we replace global updates with more local ones, so only weights responsible for the observed data sample are updated. In this work, we propose the use of dynamic self-organizing map (DSOM) with neural networks to induce such locality in the updates without ER buffers. Our method learns a DSOM to produce a mask to reweigh each hidden unit's output, modulating its degree of use. It prevents interference by replacing global updates with local ones, conditioned on the agent's state. We validate our method on standard RL benchmarks including Mountain Car and Lunar Lander, where existing methods often fail to learn without ER. Empirically, we show that our online and fully incremental method is on par with and in some cases, better than state-of-the-art in terms of final performance and learning speed. We provide visualizations and quantitative measures to show that our method indeed mitigates interference.
\end{abstract}

\section{Motivation}

Neural networks have recently achieved notable success in different areas including machine translation \citep{bahdanau2014neural} and image classification \citep{krizhevsky2012imagenet}. Despite all the successes, there remain unaddressed challenges for neural networks in order to learn efficiently. One such challenge is that of catastrophic interference: neural networks forget what they learned in the past when they are presented with new data \citep{french1999catastrophic}.

In this work, we focus on a setting in which the neural network observes each sample only once, learns from the sample, updates the weights, and then disposes of the sample without using it again in the future. Apart from its close affinity with how humans learn, this setting is popular in the linear reinforcement learning (RL) framework \citep{sutton2018reinforcement}. In the rest of this paper, we refer to this setting as online \emph{and fully incremental}, whereas we refer to settings that use experience replay (ER) buffers only as online learning frameworks. The fully incremental setting is of interest because interference tends to be most severe in such cases. Actually, the interference is so severe that the agent fails to learn how to solve a single small task successfully (\citep{ghiassian2018two}).

Neural networks inherently suffer from catastrophic interference. The network has a single set of shared weights that leads to high representational overlap across data samples \citep{french1999catastrophic}. One update can easily overwrite what was learned previously by changing too many weights, generalizing too globally. Interference is especially pronounced when the data is received in a non-independent-and-identically-distributed manner (non-i.i.d.) \citep{mccloskey1989catastrophic}, a property of the online and fully incremental setting that we consider. %Using value-based methods as an example to interpret catastrophic interference in RL, an update to a value or state-action value estimation of a particular state may potentially influence estimations of the entire state space due to the global generalization nature of neural networks. What was learned in an area of the state space can be overwritten as the agent starts learning in a different area. Such interference leads to deterioration of performance.

To mitigate catastrophic interference in RL, practitioners have resorted to approaches with ER buffers that collect recent data to perform minibatch updates \citep{volodymyr2015human}. Such minibatch updates approximate an i.i.d. data distribution which attenuates catastrophic interference. Still, as we tackle more complex problems, having an ER buffer might become infeasible in terms of memory. We cannot expect an intelligent agent to store all past experiences and to constantly relearn what it had learned in the past. Thus, it is essential to look for potential ways to achieve online RL in the absence of ER.

Many solutions have been proposed to tackle catastrophic interference. \cite{kirkpatrick2017overcoming} introduced Elastic Weight Consolidation as a regularization strategy to encourage parameters to stay as close to the parameters of previous tasks as possible. Similar to our method, \cite{DBLP:journals/corr/abs-1904-09330} proposed using self-organizing map (SOM) to mitigate interference. However, it is limited to supervised learning and the SOM used is unsuitable for cases when the data distribution is non-i.i.d.  In terms of online and fully incremental RL, \cite{ghiassian2018two} proposed using tile coding and geometric projection to sparsify input features, reducing activation overlap. But tile coding increases the number of input dimensions to a neural network, which can lead to scalability issues for benchmarks with high dimensionality. \cite{liu2018utility} introduced regularization strategies like a distributional regularizer to induce sparse representations in neural networks. The representations induced can provide locality that averts interference. Yet, these methods require pretraining and have not been extended to the online and fully incremental setting.

In this paper, we propose a method that helps an RL agent learn online and fully incrementally without the use of ER buffers and target networks. We show, on two RL tasks, that our method is on par with, and in some cases, better than baseline methods that use ER and target networks in both performance and speed. We use visualizations, and qualitative measures proposed in \cite{riemer2018learning} to show that the proposed method indeed reduces interference.

\section{Dynamic Self-Organizing Maps for Online Reinforcement learning}
As mentioned before, catastrophic interference is a byproduct of global updates to the neural network weights. The obvious solution to the problem seems to be replacing the global updates with more local ones. When an agent updates its value estimation of a state, the update should only affect value estimations of states that are similar to the current state. With this intuition in mind, we hypothesize the catastrophic interference caused by the global property of neural networks to be a major factor of neural networks' inability to learn online and fully incrementally. If we can limit the scope of an update to be more local in the state space, we may be able to achieve generalization without the interference induced by global updates.

To localize the updates and alleviate interference, we propose the use of dynamic self-organizing maps (DSOM). The proposed method learns a neural network and a DSOM in parallel. It reduces interference by inducing state-conditioned updates, removing the need for ER and target networks. Self-Organizing Maps (SOM) are a type of neural network learned without supervision \citep{kohonen1990self}. The goal of SOM is to transform an input space into a one-dimensional or two-dimensional discrete map in a topologically ordered manner, which means data samples that are close to each other in their input space would be close together on the map. Unlike standard neural networks, SOM learn with competitive learning instead of error backpropagation. Under competitive learning, other than updating all neurons, only the winner neuron gets updated. The winner is determined by a metric (e.g. euclidean norm for SOM). With SOM, both the winner neuron and its neighboring neurons on the map get updated. The neighborhood of neurons is defined by a neighborhood function. By iterating this process, the map of neurons would converge with each vector representing a cluster of the input space, conditioned on received data\textquotesingle s distribution. 

Typically, a SOM contains a set of \(N\) \(d\)-dimensional weight vectors, with each denoted as \(\vecw_i  \in \mathbb{R}^d\), \(i \in N\). Each \(\vecw_i\) is associated with a unique position \(p_i\) on a one dimensional or two dimensional grid map to represent a node. At each iteration, an input vector \(\vecv \in \mathbb{R}^d\) will be matched with the closest node \(\vecb\) on the map by 

\begin{equation} 
    \vecb = \argmin_{i\in N}(|| \vecv - \vecw_i ||),
\end{equation}

\noindent  where \(||\cdot||\) is the euclidean norm. The closest unit is commonly known as the best matching unit (BMU) or the winner neuron. The weight vector associated with the BMU and weight vectors of the BMU\textquotesingle s topologically neighboring nodes are updated to reduce the error between the input vector and those weights, pushing that neighborhood of weight vectors to be closer to the input vector. The neighborhood is determined by a neighborhood function \(h_\sigma(t, i, j)\). In this work, we use DSOM, a type of SOM proposed in \cite{rougier2011dynamic}. At each time step \(t\), the weight vectors are shifted towards \(\vecv\) by

\begin{equation}\label{eq:dsom_update}
    \Delta \vecw_i = \epsilon \cdot ||\vecv - \vecw_i||_\Omega \cdot h_\eta(i, \vecb, \vecv) \cdot (\vecv - \vecw_i) ,
\end{equation}

\noindent with the neighbourhood function defined as 

\begin{equation}\label{eq:dsom_neighborhood_func}
    h_\sigma(i, \vecb, \vecv) = e^{-\frac{1}{\eta^2}\frac{||p_i - p_b||^2}{||\vecv - \vecw_b||_\Omega^2}}.
\end{equation}
\noindent \(||\cdot||_\Omega\), \(\epsilon\) and \(\eta\) refer to the normalized euclidean norm, DSOM learning rate and elasticity (plasticity) respectively. Unlike SOM, DSOM removes the condition on time which allows the use of SOM with non-i.i.d. data without the need for an offline training period, making it a suitable candidate to work under the online and fully incremental setting. Note that elasticity modulates the coupling strength between DSOM weight vectors. If elasticity is high, weight vectors tend to be relatively close while a lower value allows looser coupling between weight vectors.

We illustrate our method with Sarsa \citep{rummery1994line} and Q-learning \citep{watkins1992q}. We parameterize the state-action value functions with neural networks for both algorithms (See Appendix \ref{app:rl_algo}).

We propose using DSOM as a resource allocation module to a neural network, modulating the extent of an update to each weight to reduce interference. At each time step, DSOM produces an output mask based on the euclidean norm between each of its weight vector and the input vector: 

\begin{equation}\label{eq:output_mask_eq}
    \Gamma = e^{-\frac{||\vecv - \vecw_i||}{\kappa}}
\end{equation}

\noindent where \(\kappa\) is a tunable hyperparameter. Weight vectors that are closer to the input vector will be weighted higher and vice versa. The number of weight vectors in DSOM is set to be the number of hidden units in the hidden layer. An element-wise multiplication is performed with the hidden layer output and DSOM\textquotesingle s output mask. By doing so, values of the hidden layer output would be modulated by the DSOM \textquotesingle s output mask based on the euclidean norm between the input vector and each DSOM weight vector. State-action values are then computed by the output layer using the modulated hidden layer output. The neural network and DSOM are learned in parallel and entirely online. See Appendix \ref{app:sam_arch} for a sample architecture. 

To put it in the context of RL, DSOM\textquotesingle s weight vectors learns to represent different parts of the state space. Each weight vector is associated with a hidden unit in the hidden layer. The output mask determines the degree of use of each hidden unit based on the state feature vector\textquotesingle s euclidean distance to each DSOM weight vector. The weighted mask has a similar effect as activation sharpening proposed by \cite{french1992semi}, except the selection of nodes to be sharpened or dampened is conditioned on the current state. With the state similarity information embedded in the output mask, an update to the state-action value estimation of a state will only affect the value estimations of similar states. Weights that are used by dissimilar states more would be weighted less in the DSOM mask, diminishing changes to those weights during an update. In other words, the learning progress of various parts of the state space cannot interfere with each other as much, facilitating the possibility for the agent to learn online and fully incrementally by having more local updates. We hypothesize the output mask from DSOM alleviates catastrophic interference by masking out interfering updates while allowing generalization across similar states. 

%Note that the weighted mask allows overlapping state aggregation like tile coding \citep{albus1975data}, not disjoint representations of the state space because a state can be close to multiple DSOM weight vectors.

\section{Experiments and Results}

We evaluated our proposed method on two RL benchmarks, namely Mountain Car and Lunar Lander (See Appendix \ref{app:env_descriptions}). The agent takes in a  feature vector as input and produces state-action values of all actions. All methods used one single hidden layer of 800 units unless specified otherwise. For our method, we ensure the total number of weights, including the number of DSOM vectors, to be the same or less than the baselines for fair comparison \footnote{For instance, to compare with a neural network of a single hidden layer of 800 units (number of weights = number of features \(\cdot\) 800 + 800 \(\cdot\) number of actions), we used 400 hidden units and 400 DSOM weight vectors for our proposed approach (number of weights = (400 + number of DSOM weight vectors) \(\cdot\) number of features + 400 \(\cdot\) number of actions). Note that the number of weights used in the latter case is less than the baseline}. The performance measure of Mountain Car is the number of steps taken in an episode with a cutoff of 1000 steps. The lower the number of steps means the less time an agent needs to solve the problem. For Lunar Lander, the episodic reward is used as the performance measure. We evaluated our method against several baselines, including baselines that use ER, target networks and adaptive learning rate (ALR) mechanisms and baselines that learn only with the RL algorithms (See Appendix \ref{app:exp_details})\footnote{Methods that do not use adaptive learning rate optimizers use stochastic gradient descent (SGD) by default}. We performed parameters sweeps over each method and report the best performance averaged over 30 runs. We used Sarsa and Q-learning in Mountain Car and Lunar Lander respectively.

\begin{figure*}[t]
    \begin{center}
       \subfigure[]{\includegraphics[scale = 0.27]{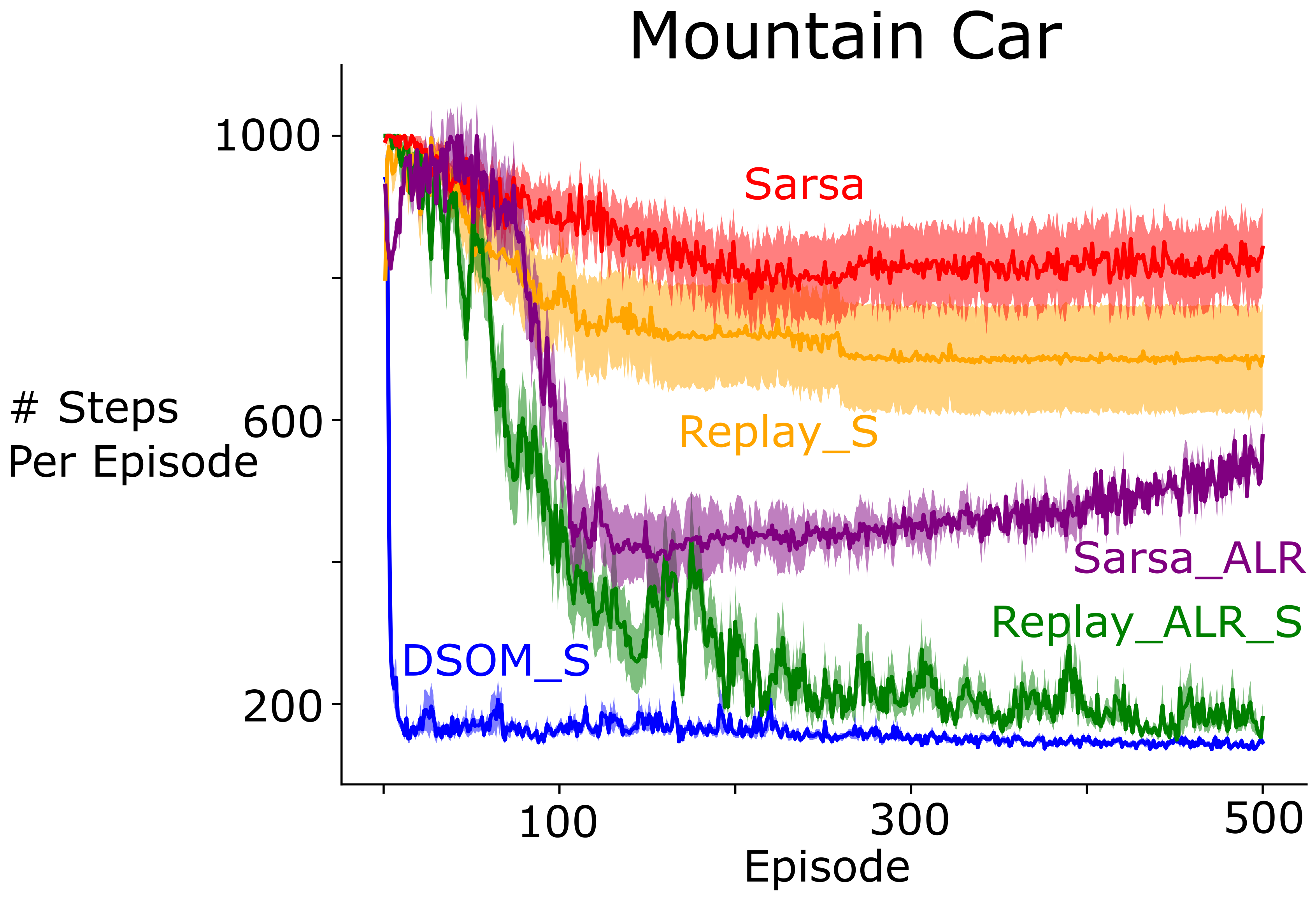}} 
        \subfigure[]{\includegraphics[scale = 0.26]{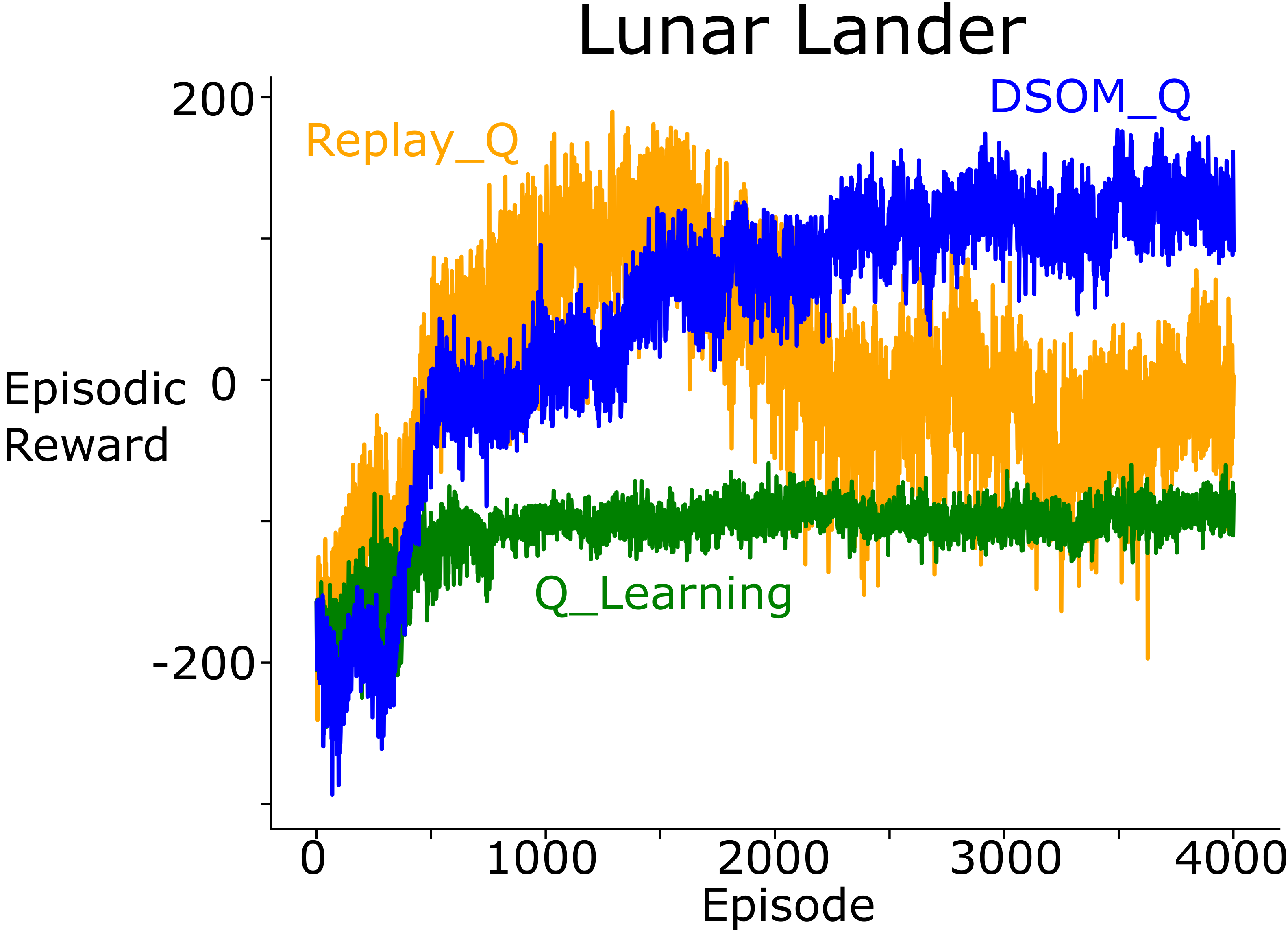}}
        \subfigure[]{\includegraphics[scale = 0.26]{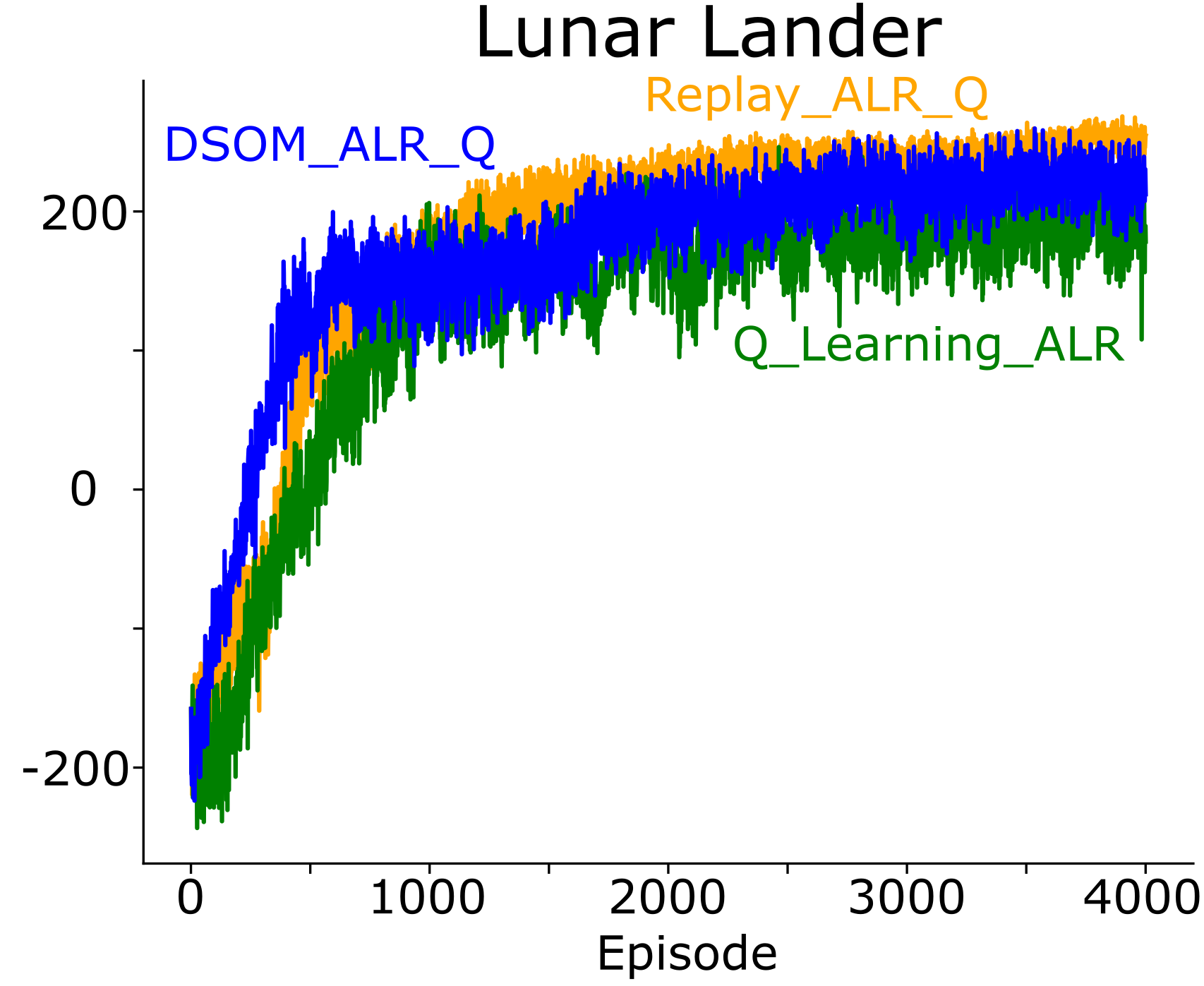}}
        \caption{Performances of algorithms on environments Mountain Car and Lunar Lander. On Mountain Car, DSOM outperforms methods with ER and target networks by mitigating interference. On Lunar Lander, DSOM achieves comparable results (with greater speed) as methods with ER and target networks. Without ALR, our method outperforms replay methods.}
        \label{fig:learning_curves}
    \end{center}
\end{figure*}

We show learning curves in Figure \ref{fig:learning_curves}. On Mountain Car, our method (DSOM\_S), which learns online without ER, outperforms the baseline methods in terms of performance and speed. For baselines without ER and target networks (Sarsa and Sarsa\_ALR), they either have trouble in solving the task or learn very slowly, likely due to catastrophic interference when updating the neural networks as pointed out in \cite{ghiassian2018two}. When ER, target network and ALR optimizer are used, the baseline (Replay\_ALR\_S) is able to learn as ER helps to alleviate some amount of interference. Our method performs well by balancing between generalization and interference directly when updating the value function. This is done by modulating the updates based on the state similarity information embedded in the DSOM output mask, affecting value estimations of dissimilar states less. This means updates are more local, influencing only the surrounding area of the current state. This achieves online learning without any explicit storage of observed data.

\begin{figure*}[t]
    \begin{center}
        \subfigure[]{\includegraphics[scale = 0.018]{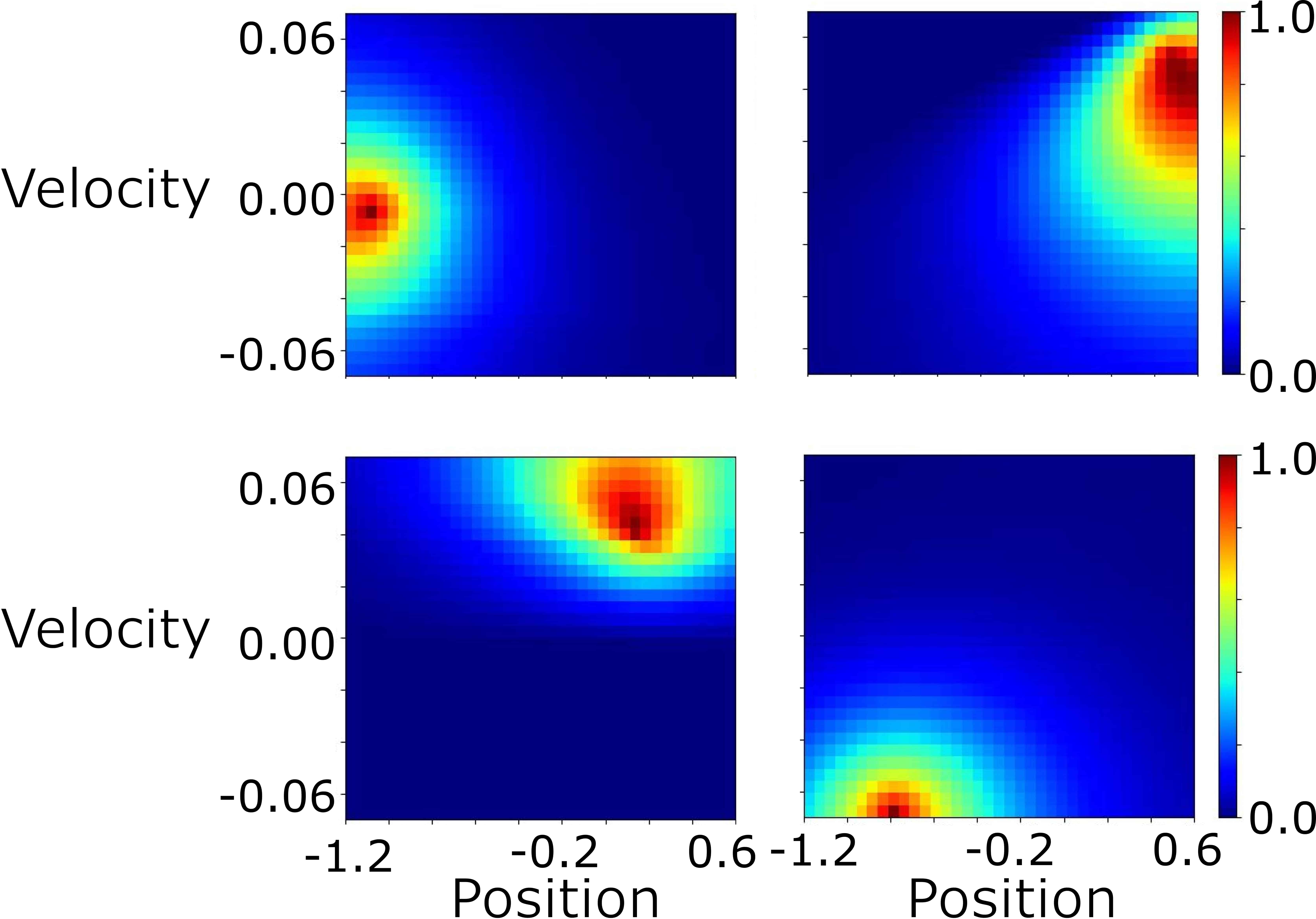}}
        \subfigure[]{\includegraphics[scale = 0.018]{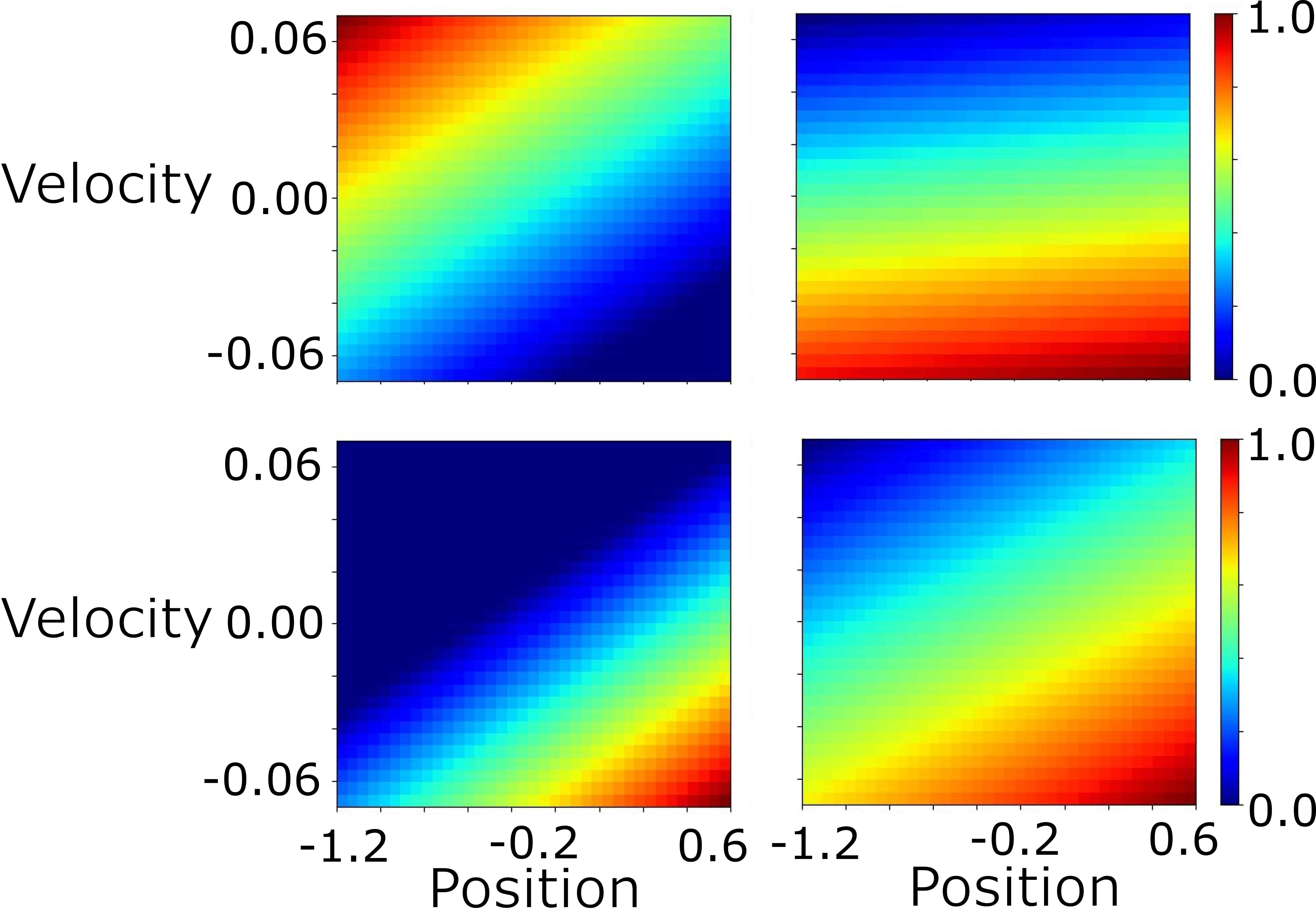}}
        \subfigure[]{\includegraphics[scale=0.27]{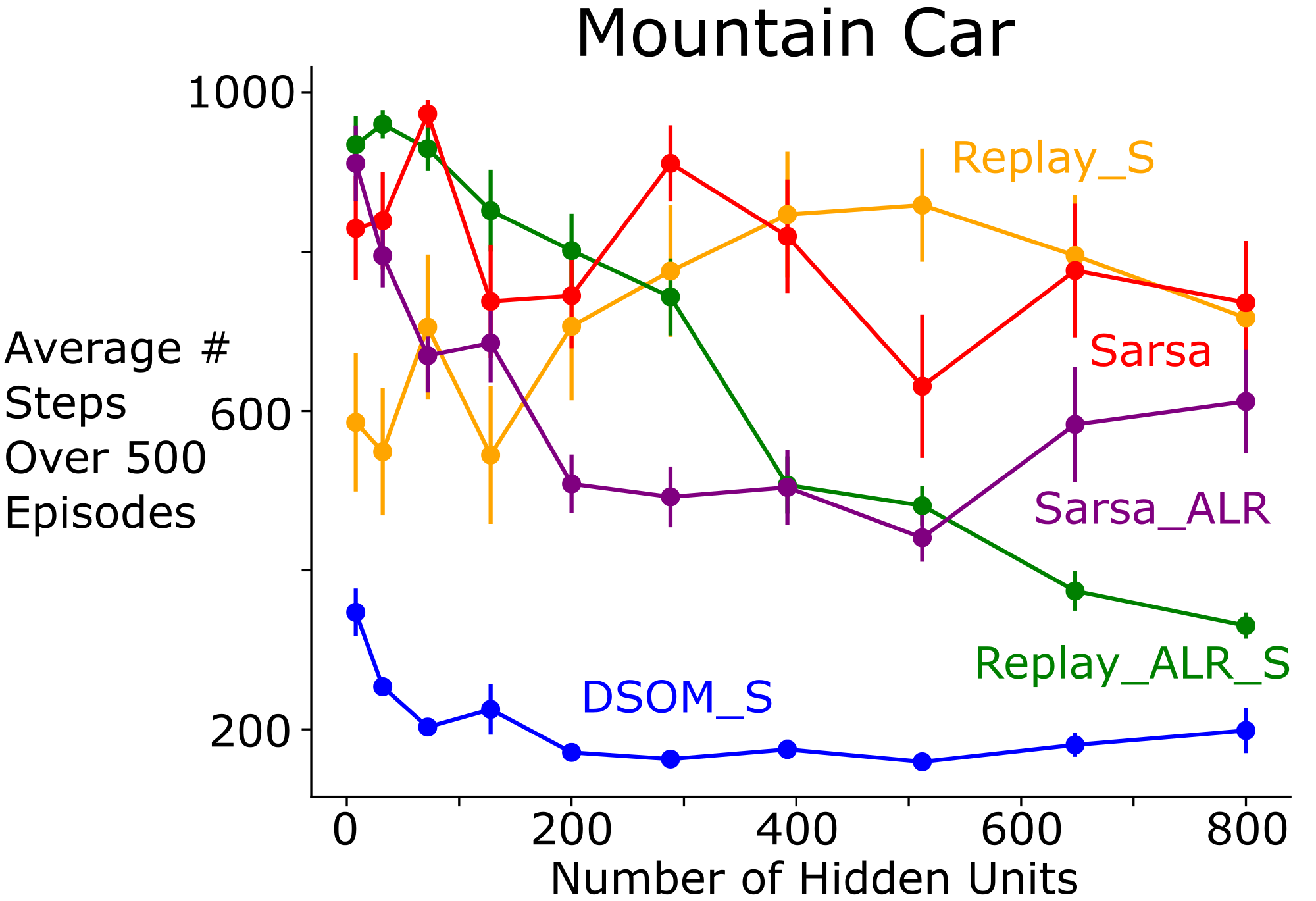}}
        \caption{(a) and (b) show heatmap visualizations of activation strength of DSOM\_S and Replay\_ALR\_S across the state space of Mountain Car. The use of DSOM enables more local and specialized activations. (c) shows performances of algorithms on Mountain Car with varying number of hidden units. The use of DSOM reduces the number of hidden units needed, solving the problems with very small number of hidden units.}
        \label{fig:heatmaps}
    \end{center}
\end{figure*}

On Lunar Lander, a more complex control problem, our method with DSOM performs as well as, and in the case of not using ALR, better than methods that use ER and target network. It also learns faster than the baselines. This further illustrates the potential benefits of online and fully incremental learning with neural networks, if we are able to do more local and differentiated (state-conditioned) updates. Note that the use of ALR optimizer appears to play a role in reducing interference. In Figure \ref{fig:learning_curves}c and \ref{fig:learning_curves}d, by removing the use of ALR optimizer, all methods perform worse. Especially in the case with ER and target network (Replay), learning instability is observed with a significant dip in performance while our method continues to improve and perform well consistently. 

\begin{table}[h]
    \centering
    \caption{Quantitative measure of interference across the methods in Mountain Car}
    \begin{tabular}{c|c|c|c|c}
        \hline
       \textbf{Online} & \textbf{Online\textunderscore ALR} & \textbf{Replay} & \textbf{Replay\textunderscore ALR} & \textbf{DSOM}\\
       \hline
        0.1739 & 0.2735 & 0.4164 & 0.1693 & \textbf{0.1079} \\
        \hline
    \end{tabular}
    \label{tab:grad_inter_table}
\end{table}

We visualize activation response functions of hidden units across the state space of Mountain Car as shown in Figure \ref{fig:heatmaps}a and \ref{fig:heatmaps}b. A set of states representative of the whole state space is used to measure the activation responses of each hidden unit (See Appendix \ref{app:quant_cf}). Based on the visualizations, hidden units of our method empirically demonstrates more local and specialized responses to specific areas of the state space. On the other hand, hidden units trained with ER, target network and ALR optimizer respond to a large area of the state space. This provides a potential explanation to the effectiveness of our method as each hidden unit only responds to a small subset of similar states. This aids the network to generalize across similar states while avoiding overwriting estimations of dissimilar states. Furthermore, we employ a quantitative measure mentioned in \cite{riemer2018learning} to quantify interference (See Appendix \ref{app:quant_cf}). This measure would be zero when there is no activation overlapping across two samples. Thus, the lower the value, the less the amount of interference is. We used this measure for each method averaging over all unique pairs of states used in the visualizations. As shown in Table \ref{tab:grad_inter_table}, our method has the least level of interference across the state space. With both qualitative and quantitative results, we provide evidence that indicates our method performs well in the online and fully incremental setting without ER by reducing interference by inducing more local updates. 

%Additionally, we found evidence showing that our method uses resources more efficiently (see Appendix \ref{app:mem_sweep}). In short, our method is able to solve the same problem with less number of hidden units (as few as 36 units in Mountain Car) because each unit is specialized in one small area of the state space. 

Additionally, with our method inducing more local updates, we hypothesize that this may lead to more efficient use of resources (hidden units) as each unit is specialized in a small area of the state space. This means our method may need less hidden units to solve the same problem. To verify this hypothesis, we conducted experiments on Mountain Car, varying the number of hidden units used. We show the results in Figure \ref{fig:heatmaps}c. For each method, we measure the average number of steps used over the number of training episodes. As we can see, our method continues to solve the tasks and outperform all the baselines used, even in cases with a very small number of hidden units \footnote{On the x-axis of Figure 3, the number of units refers to the total number of hidden units in the hidden layer for the baselines. For our method, it refers to the total number of DSOM weight vectors + the total number of hidden units, as they are of the same feature length.}. Our method is able to solve Mountain Car with as few as 36 hidden units. The results support the hypothesis that our method's capability to reduce interference may give rise to more efficient use of resources in neural networks, reducing the number of hidden units needed.

\section{Conclusion}
We proposed a method that combines dynamic self-organizing maps with neural networks to solve reinforcement learning problems fully incrementally, a setting akin to how humans learn. The method achieves fully incremental learning by localizing the updates and preventing interference with what the network has learned in the past. It removes the need for experience replay buffers and target networks. It also provides a new perspective on how interference can be avoided in neural networks.

\subsubsection*{Acknowledgments}
The authors thank Richard Sutton and Khurram Javed for discussions contributing to the results presented in this paper. The authors gratefully acknowledge funding from, Alberta Machine Intelligence Institute, JPMorgan Chase \& Co, the Natural Sciences and Engineering Research Council of Canada, and Google DeepMind.

\bibliographystyle{plainnat}
\bibliography{references}

\section{Appendix}

\subsection{RL Algorithms with neural networks}
\label{app:rl_algo}
In reinforcement learning, an agent interacts with its environment by taking actions at discrete time steps $t=0, 1, 2, \cdots$. The environment is commonly formulated as a Markov Decision Process (MDP) with states \(\mathcal{S}\), actions \(\mathcal{A}\), transition probabilities \(P : \mathcal{S \times A \times S} \rightarrow [0, 1]\), rewards \(\mathcal{R : S \times A \times S \rightarrow R}\) and discount function \(\mathcal{\gamma : S \times A \times S} \rightarrow [0, 1]\) \cite{white2017unifying}. At each time step $t$, the agent is in a state $S_t$, and takes an action $A_t$. In response, the environment emits a reward $R_{t+1}$ and takes the agent to a state $S_{t+1}$. The goal of the agent is to maximize the return, defined as the discounted sum of the cumulative rewards: 

\begin{equation}
   G_t ={\sum_{k=0}^{\infty} {\gamma^k R_{t+k+1}} }, 
\end{equation}

In this paper, we use Sarsa \cite{rummery1994line} and Q-learning \cite{watkins1992q} algorithms to test our method in control problems. In both algorithms, the agent learns to approximate the state-action value function and acts near-greedily according to those state-action values. The state-action values for a policy \(\pi : \mathcal{S} \times \mathcal{A} \rightarrow [0, 1]\) are the expected return for that policy beginning from state \(s\) and action \(a\): 

\begin{equation}
    q_\pi(s, a) = \mathbb{E}_\pi[G_t | S_t = s, A_t = a].
\end{equation}
\noindent where \(\mathbb{E}_\pi[\cdot]\) denotes taking expectation under policy \(\pi\).

We parameterize the state-action value functions with neural networks for Sarsa and Q-learning, denoted as \(\hat{q}(S_t, A_t, w_t)\), where \(w_t\) refers to the neural network weights. The state-action value function is commonly learned by bootstrapping state-action value of the next state, minimizing the temporal difference error. The update rules for Sarsa and Q-learning to learn the state-action value function are as follows:

\begin{equation}
     w_{t+1} = w_t + \alpha(R_{t+1} +  \gamma\hat{q}(S_{t+1}, A_{t+1})  - \hat{q}(S_t, A_t))\nabla_w \hat{q}(S_t, A_t), 
\end{equation}

\begin{equation}
    w_{t+1} = w_t + \alpha(R_{t+1} + \gamma\max_{A_{t+1}}\hat{q}(S_{t+1}, A_{t+1}) - \hat{q}(S_t, A_t))\nabla_w \hat{q}(S_t, A_t), 
\end{equation}

\noindent where \(\alpha\) and \(\gamma\) learning rate and discount factor respectively. Each neural network takes in a state feature vector as an input and produces state-action values for each possible action. All neural networks used in this work use ReLU as activation function and a linear output layer.

\subsection{Sample Architecture of using DSOM}
\label{app:sam_arch}
\begin{figure}[h]
    \begin{center}
        \includegraphics[scale=0.2]{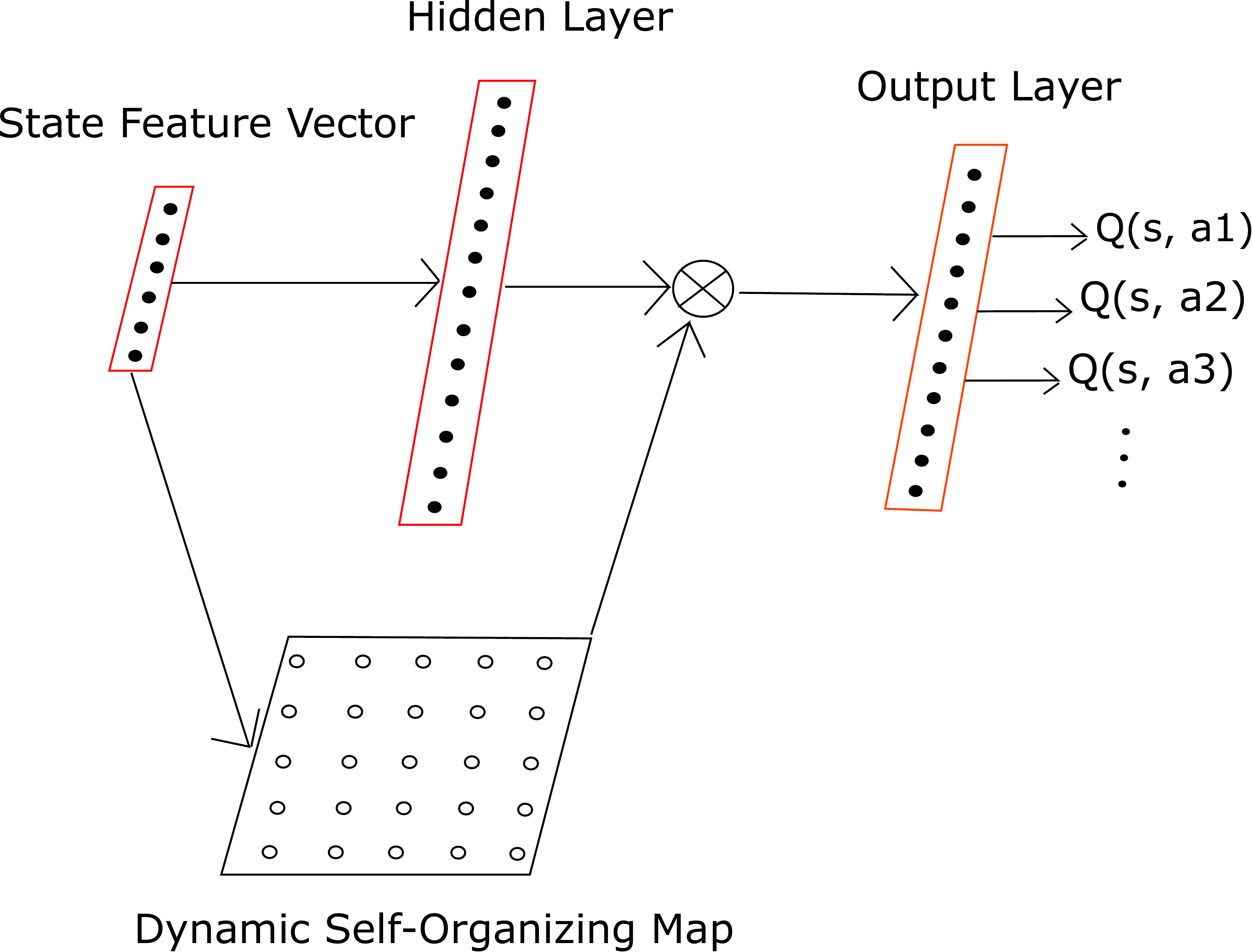}
    \end{center}
    \caption{Sample Architecture of DSOM-DQN}
    \label{fig:sam_arch}
\end{figure}

\subsection{Details on heatmap visualizations}
\label{app:heatmap_vis}
we used a set \(S\) of 121 states that covers the entire state space. These states were generated in the manner as follows: \(S = {(-1.2 + 0.17x, -0.07 + 0.014y) : x, y = 0, 1, 2, ..., 10}\). For each algorithm, the normalized activation values of the hidden layer for states in \(S\) were used to create the heatmaps 

\subsection{Quantitative measure of Interference}
\label{app:quant_cf}
Introduced in \cite{riemer2018learning}, The measure looks at the dot product of the gradient vector of two samples with respect to the parameters: 

\begin{equation}\label{eq:grad_inter_eq}
    \frac{\partial L(x_1, y_1)}{\partial \theta} \cdot \frac{\partial L(x_2, y_2)}{\partial \theta}
\end{equation}

\noindent where L corresponds to the loss function.

\subsection{Environment Descriptions}
\label{app:env_descriptions}
\textbf{Mountain Car}: The Mountain Car environment has a 2-dimensional state space: position and velocity. The value of position is between -1.2 and 0.6 and the value of velocity is between -0.07 and 0.07. The agent has three discrete actions, namely full throttle forward, full-throttle backward and no throttle. The car is initialized around the hill bottom randomly with a reward of -1 for each time step before it gets to the top of the hill. The goal state (top of the hill) is defined to be when the position is greater than 0.5. The episode terminates if it takes more than 1000 steps. 

\noindent \textbf{Lunar Lander}: The Lunar Lander environment has an 8-dimensional state space: x-coordinate, y-coordinate, x-velocity, y-velocity, angle, angular velocity, and two binary features to indicate whether the left and right legs of the lander are in contact with the ground. The agent has four discrete actions, namely, fire left orientation engine, fire right orientation engine, do nothing and fire main engine. we used the Lunar Lander environment from Open AI Gym \cite{brockman2016openai}. The goal of an agent is to land the lander safely on a fixed launching pad without crashing. The reward for landing on the pad with a zero resultant speed is between 100 and 140 with a negative reward if it moves away from the pad. A reward of 10 is given for each leg's ground contact and A punishment of -0.3 is given for firing the main engine at every frame. The episode terminates either when the lander crashes or it comes to rest.

\subsection{Experimental Details}
\label{app:exp_details}
For Mountain Car, we used Sarsa as the RL algorithm. We evaluated 5 different methods:

\begin{itemize}
    \item Sarsa\footnote{Methods that do not use adaptive learning rate optimizers use stochastic gradient descent (SGD) by default},
    \item Sarsa with ALR optimizer, abbreviated as Sarsa\_ALR in the figures,
    \item Sarsa with DSOM, abbreviated as DSOM\_S.
    \item Sarsa with experience replay buffer and target network, abbreviated as Replay\_S and
    \item Sarsa with experience replay buffer, target network and ALR optimizer, abbreviated as Replay\_ALR\_S.
\end{itemize}

For Lunar Lander, we used Q-learning as the RL algorithm. We evaluated 6 different methods:

\begin{itemize}
    \item Q-learning,
    \item Q-learning with ALR optimizer, abbreviated as Q\_Learning\_ALR,
    \item Q-learning with DSOM, abbreviated as DSOM\_Q,
    \item Q-learning with DSOM and ALR optimizer, abbreviated as DSOM\_ALR\_Q,
    \item Q-learning with experience replay buffer and target network, abbreviated as Replay\_Q and 
    \item Q-learning with experience replay buffer, target network and ALR optimizer, abbreviated as Replay\_ALR\_Q.
\end{itemize}

In terms of exploration policy, we used the same exploration policy for all methods in each environment. For both Mountain Car, we used an \(\epsilon\)-greedy policy to take a random action with a probability of 10\%. For Lunar Lander, we used a decaying \(\epsilon\)-greedy policy. The Starting \(\epsilon\), ending \(\epsilon\) and \(\epsilon\) decay rate are 1.0, 0.1 and 0.995 respectively. Here we present the hyperparameter ranges that were swept over for the experimental runs in each environment. 

\begin{table}[h]
\begin{tabular}{lll}
\cline{1-3}
\multicolumn{1}{c}{} & Mountain Car & Lunar Lander \\ \hline
 & \textbf{Sarsa \& Sarsa\_ALR} & \textbf{Q\_Learning \& Q\_Learning\_ALR} \\ \cline{1-3}
Learning Rate & \(0.005^c, c \in {0, 1, 2, 3, 4}\) & \(0.05^c, c \in {0, 1, 2, 3, 4}\) \\
Number of Hidden Units & \(2 \cdot n^2, n \in {2, 4, 6, .., 18, 20}\) & 800 \\
Optimizer & RMSProp/SGD & Adam/SGD \\ \cline{1-3}
%\multicolumn{4}{l}{\textbf{Replay and Replay\_ALR}} \\ \cline{1-4}
 & \textbf{Replay\_S \& Replay\_ALR\_S} & \textbf{Replay\_Q \& Replay\_ALR\_Q} \\ \cline{1-3}
Learning Rate & \(0.005^c, c \in {0, 1, 2, 3, 4}\) & \(0.05^c, c \in {0, 1, 2, 3, 4}\)  \\ 
Number of Hidden Units & \(2 \cdot n^2, n \in {2, 4, 6, .., 18, 20}\) & 800  \\
Optimizer & RMSProp/SGD & Adam/SGD \\
Replay Size & 20000 & 100000 \\
Batch Size & 32 & 64 \\
Target Network Update Frequency & 10 & N/A \\
Target Network Soft Update Frequency & N/A & 4 \\
Target Network Soft Update Ratio \(\tau\) & N/A & \(0.01^c, c \in {-1, 0, 1, 2, 3}\) \\ \cline{1-3}
%\multicolumn{4}{l}{\textbf{DSOM}} \\ \cline{1-4}
 & \textbf{DSOM\_S} & \textbf{DSOM\_Q \& DSOM\_ALR\_Q} \\ \cline{1-3}
Learning Rate & \(0.005^c, c \in {0, 1, 2, 3, 4}\) & \(0.05^c, c \in {0, 1, 2, 3, 4}\)  \\
DSOM Learning rate \(\epsilon\) & \(2^c, c \in {-1, -2, -3, -4}\) & \(2^c, c \in {-1, -2, -3, -4}\) \\
Optimizer & RMSProp/SGD & Adam/SGD \\
Elasticity \(\eta\) & \(2^c, c \in {-2, -1, 0, 1, 2, 3, 4}\) & \(2^c, c \in {-2, -1, 0, 1, 2, 3, 4}\) \\
\(\kappa\) & 0.5 & 0.5 \\
Number of DSOM Weight Vectors & \(m^2, m \in {2, 4, 6, .., 18, 20}\) & 400 \\
\end{tabular}
\end{table}

\end{document}